\documentclass[letterpaper, 10 pt, conference]{ieeeconf}  

\IEEEoverridecommandlockouts                              

\usepackage{hyperref}
\usepackage{algorithm}
\usepackage{graphicx}
\usepackage{algpseudocode} 
\usepackage{amsmath} 
\usepackage{amsfonts}
\usepackage{cleveref}
\usepackage[inline]{enumitem}
\usepackage{float}
\usepackage{subfig}
\usepackage{xcolor}
\usepackage[export]{adjustbox}
\usepackage{booktabs}

\title{\LARGE \bf
Multi-Agent Transfer Learning via Temporal Contrastive Learning
}

\author{
    Weihao Zeng, Joseph Campbell, Simon Stepputtis, Katia Sycara \\
    Carnegie Mellon University
}

\DeclareMathOperator{\E}{\mathbb{E}}

\begin{document}

\maketitle
\thispagestyle{empty}
\pagestyle{empty}

\begin{abstract}

This paper introduces a novel transfer learning framework for deep multi-agent reinforcement learning. The approach automatically combines goal-conditioned policies with temporal contrastive learning to discover meaningful sub-goals. The approach involves pre-training a goal-conditioned agent, finetuning it on the target domain, and using contrastive learning to construct a planning graph that guides the agent via sub-goals. Experiments on multi-agent coordination Overcooked tasks demonstrate improved sample efficiency, the ability to solve sparse-reward and long-horizon problems, and enhanced interpretability compared to baselines. The results highlight the effectiveness of integrating goal-conditioned policies with unsupervised temporal abstraction learning for complex multi-agent transfer learning.
Compared to state-of-the-art baselines, our method achieves the same or better performances while requiring only 21.7\% of the training samples.

\end{abstract}

\section{Introduction}

\begin{figure*}[htbp!]
    \centering
    \includegraphics[width=0.9\textwidth]{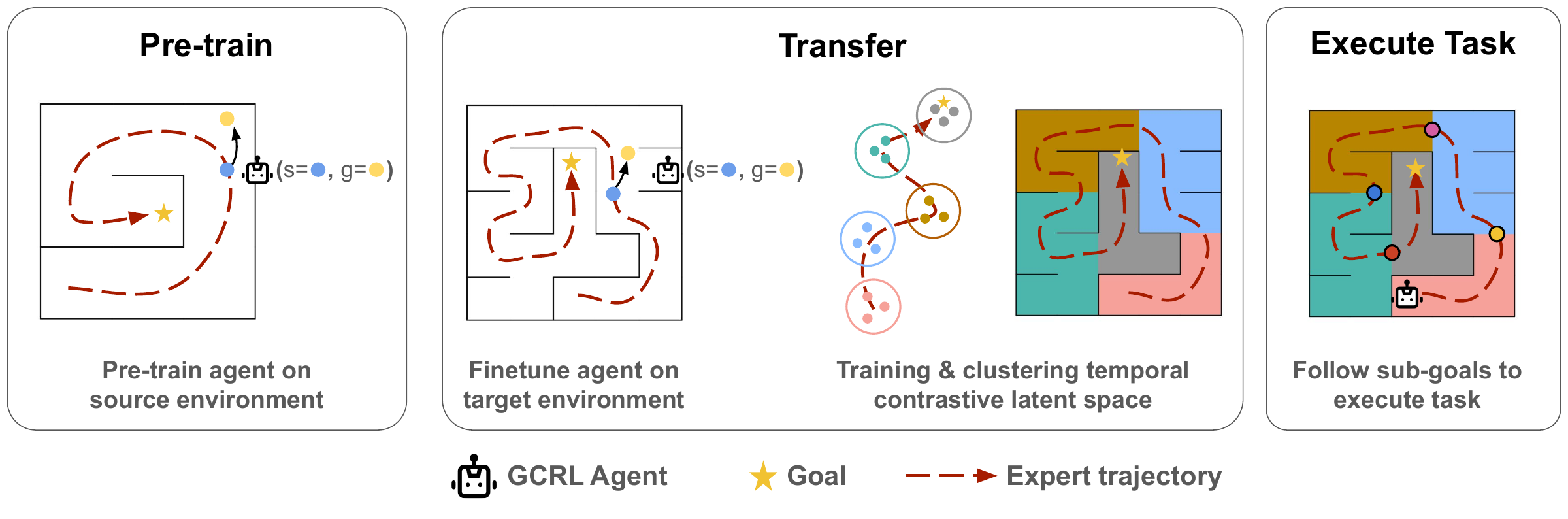}
    \caption{ Our method follows three steps: 1) pre-train the GCRL agent to acquire diverse transferable skills by achieving short-horizon goals in the source environment; 2) finetune the GCRL agent on the target environment, learn a latent space to encapsulate the temporal structure of trajectories form rolling out the finetuned GCRL agent, and construct a planning graph, whose nodes are clusters from the latent space and edges are transitions between clusters observed in the expert trajectory; 3) and, execute task in the target environment by following sub-goals.
    We assume a single successful demonstration in the target environment is given, which we utilize to guide agent finetuning and graph construction.
    }
    \label{fig:method_summary}
\end{figure*}

Reinforcement Learning (RL) has emerged as a powerful framework for solving decision-making problems by enabling agents to learn optimal policies through interactions with the environments. It has achieved remarkable success in various challenging domains, such as robotics \cite{lu2021aw, ma2023eureka, andrychowicz2020learning} and game-playing \cite{vinyals2017starcraft, silver2017mastering}.
Despite its impressive advancements, RL often struggles with sample inefficiency, requiring many interactions with the environment to learn effective policies \cite{zhu2023transfer}.
This problem is exacerbated in multi-agent systems, where the size of the state and action space increase combinatorially with the number of agents. Moreover, sparse rewards and partial observability can further worsen the sample inefficiency of RL algorithms \cite{zhu2023transfer}.

Transfer learning (TL) has emerged as a promising approach to address these challenges and improve the sample efficiency of RL. TL aims to leverage knowledge learned in a task to accelerate learning and boost performance in a target task \cite{zhu2023transfer}. The key idea behind TL is that related tasks often share common structures or features, which can be extracted and reused instead of learned from scratch.
For example, skills learned in previous navigation tasks can transfer to new navigation goals or environment layouts \cite{zhang2017deep}.

This paper introduces a novel transfer learning framework that integrates goal-conditioned reinforcement learning (GCRL) policies \cite{schaul2015universal} with unsupervised temporal abstraction learning and graph-based planning to capture and exploit reusable knowledge across tasks. Our approach employs contrastive learning \cite{chen2020simple} to learn a compact representation of the temporal structure from agent trajectories and then transforms this learned latent space into a graph through clustering. The resulting graph encodes abstract states as nodes representing clusters of similar states and temporal transitions between these clusters as edges. This graph structure enables efficient planning and sub-goal generation, guiding the GCRL policy in the target domain. Our approach consists of three main steps:
\begin{enumerate*}
\item First, we train a GCRL agent by reaching diverse short-horizon goals in the source domain, enabling it to acquire diverse skills for reaching various goals.
\item Next, we finetune the GCRL agent on the target domain, learn a latent space of the temporal structure from the trajectories generated by the GCRL agent using contrastive learning \cite{chen2020simple}, and construct a planning graph from the latent space.
\item Finally, we guide the GCRL agent using sub-goals generated from the planning graph to complete the task in the target domain.
\end{enumerate*}

We demonstrate the effectiveness of our proposed framework through extensive experiments across multiple multi-agent transfer scenarios on the Overcooked environment \cite{carroll2019utility}. Our approach offers several key benefits, including:
\begin{enumerate*}
    \item improved sample efficiency when learning new tasks,
    \item the ability to solve challenging sparse-reward or long-horizon tasks by leveraging the learned temporal abstractions and
    \item enhanced the learning process's interpretability by discovering meaningful sub-goals and skills.
\end{enumerate*}

The main contributions of this paper are:
\begin{enumerate}
    \item We introduced a novel TL approach for RL that enables agents to learn new tasks efficiently by leveraging prior experience.
    \item We combined goal-conditioned policies with unsupervised learning of temporal abstractions, enabling more sample-efficient and adaptable RL agents.
\end{enumerate}

\section{Related Works}

\textbf{GCRL} agents learn policies for achieving specified goal states \cite{schaul2015universal, andrychowicz2017hindsight} instead of performing fixed tasks. Recent works have extended GCRL to multi-goal scenarios \cite{plappert2018multi}, hierarchical goal-setting \cite{nachum2018data,levy2017learning}, and exploration in sparse reward settings \cite{ecoffet2019go, pong2019skew, li2021genurl, sekar2020planning, mendonca2021discovering, hu2023planning}.

\textbf{Contrastive learning} has been applied to robotics for learning state and reward representations, improving sample efficiency and generalization in control tasks \cite{laskin2020curl, zhan2022learning}. It has also been used for learning invariant representations \cite{laskin2020reinforcement}, view-angle invariant representations \cite{florence2018dense, cao2022reinforcement}, and sim-to-real transfer \cite{cao2023learning}. Recent works have used contrastive learning to preserve temporal structure in latent representations \cite{park2024foundation}.

\textbf{Hierarchical reinforcement learning} (HRL) learns a hierarchy of policies at different abstraction levels to solve complex tasks efficiently \cite{sutton1999between, bacon2017option}. Recent works have explored goal-conditioned hierarchical policies \cite{nachum2018data,levy2017learning} and combined HRL with meta-learning \cite{frans2017meta}.

\textbf{Transfer learning} in RL leverages knowledge from source tasks to improve learning efficiency and performance in target tasks \cite{zhu2023transfer}. Approaches include Progressive Neural Networks \cite{rusu2016progressive}, learning invariant feature spaces \cite{gupta2017learning}, meta-learning \cite{finn2017model}, learning transferable representations \cite{higgins2017darla}, policy distillation \cite{czarnecki2019distilling, schmitt2018kickstarting}, and curriculum learning \cite{uchendu2023jump}.



\section{Method}

Our transfer learning framework facilitates the efficient adaptation of trained agents to new environments through a three-stage approach, as shown in \autoref{fig:method_summary}:
\begin{enumerate*}
\item training a GCRL agent on a source environment to acquire diverse skills that can be leveraged in target environments, as shown in \autoref{sec:method_train_gcrl};
\item finetuning the GCRL agent on the target environment, learning a latent representation of the agent's behavior using contrastive learning to capture the temporal structure of the agent's trajectories, and constructing a planning graph based on the learned latent space, as shown in \autoref{sec:method_temp_cluster};
and \item execute the task in the target environment by following sub-goals generated from the planning graph using the finetuned GCRL agent as shown in \autoref{sec:method_transfer}.
\end{enumerate*}
We assume access to a single demonstration of successful task completion in the target environment, which we utilize to guide agent finetuning and graph construction.
During training the GCRL agent, by resetting to states in the expert trajectory, we allow the GCRL agent to focus on state regions related to completing the task rather than searching over a much larger space for finding the optimal policy, which improves sample efficiency \cite{vemula2023virtues}.


\subsection{Goal-Conditioned Reinforcement Learning Agent}
\label{sec:method_train_gcrl}

To train the GCRL agents, we utilize the universal value approximator \cite{schaul2015universal} and Proximal Policy Optimization \cite{schulman2017proximal}.
We assume we can sample goal states for a given initial state. On each episode, we sample the initial state from the expert trajectory $\tau_{\text{expert}}$ and sample a goal state $g \sim P(s_0 | g)$, where $P(s_0 | g)$ is sampling a goal state by random walking from $s_0$.
For a comprehensive algorithm description, we refer the reader to \cite{schaul2015universal} and \cite{schulman2017proximal}.
Upon transferring, we first finetune the GCRL agent on the target environment and perform temporal contrastive learning and clustering.

\subsection{Temporal Contrastive Learning and Clustering}
\label{sec:method_temp_cluster}

Providing sub-goals guiding the GCRL agents to complete tasks in target environments is a promising avenue to efficiently transfer skills learned in the source environment to the target environment. This motivates the efficient construction of planning graphs grounded in agent behaviors.
To achieve this, we utilize contrastive learning to distill a latent space representing temporal distances, specifically, the minimal steps required for an agent to transition from one state to another.
However, obtaining the minimal temporal distance between state pairs is hard because this requires optimal control between every pair of states. Hence, we use state pairs and corresponding temporal distances from rollouts generated by the GCRL agent for approximation. The resulting temporal distances are noisy.
Hence, we employ the InfoNCE \cite{oord2018representation} approach to learn a mapping $f_w$ from the observational space to the embedding space, where geometric proximities in the embedding space mirror temporal distances in the trajectories.
This relationship is encapsulated in \autoref{eq:temporal_contrastive}, with $d(\cdot, \cdot)$ representing a metric distance function. We choose $d(\cdot, \cdot)$ as the L2 distance in this paper.
Adopting a metric space as $d(\cdot, \cdot)$ enables estimating temporal distances between unobserved state pairs using the triangular inequality.
This contrastive learning and metric formulation, coupled with neural network modeling, empowers our system to process and generalize from noisy trajectory data.
During training, we select state pairs within $T$ timesteps in a trajectory to be positive samples and randomly sample states within the same batch to be negative samples. $T$ is a hyper-parameter governing the maximum temporal threshold for positive sample pairs.



\begin{equation}
\label{eq:temporal_contrastive}
\resizebox{0.9\linewidth}{!}{$
    L_{\text{tc}}(x, x_{pos}, X) = - \E \left[ \log \frac{exp(-d(f_w(x), f_w(x_{pos})))}{ \sum_{x' \in X} exp(-d(f_w(x), f_w(x'))) } \right]
$}
\end{equation}

Note that the learned latent space reflects the temporal distances of the underlying trajectories used for training.
Thus, curating a dataset representative of the state and transition distribution for the designated task is crucial.
Collecting rollouts of states relevant to the desired task with temporal distances close to the minimal temporal distances is essential for learning latent space structures useful for the task.

In \autoref{alg:train_tempotal_latent}, we sample initial states from an expert trajectory $\tau_{\text{expert}}$ to ensure we efficiently cover state regions relevant to the completing the task; we use the trained GCRL agent $\pi_\theta$ to collect rollouts; furthermore, we sample state pairs to balance the probabilities of sampling each state. After training $f_w$, we fit a cluster classifier to the latent features $f_w(\text{Dataset})$. Finally, we construct a graph where the nodes are the clusters and edges are the cluster transitions from the expert trajectory.

\begin{algorithm}[htp!]
\caption{Training Temporal Latent Space}
\label{alg:train_tempotal_latent}
\begin{algorithmic}[1]
\State \textbf{Input:} env, $f_w$, $\pi_\theta$, $\tau_{\text{expert}}$, $P(s_0 | g)$
\State $s_0 \sim \tau_{\text{expert}}$
\State $g \sim P(g|s_0)$ 
\State Dataset $\gets$ rollouts($\pi_\theta$, env, $s_0$, $g$)
\While{not converged}
    \State $x, x_{pos}, X \gets$ BalancedSampling(Dataset)
    \State Optimize $L_{\text{tc}}(x, x_{pos}, X)$
\EndWhile
\State ClusterClassifier $\gets$ Cluster $f_w(\text{Dataset})$
\State PlanningGraph $\gets$ construct\_graph(Dataset, $f_w$, $\tau_{\text{expert}}$)
\end{algorithmic}
\end{algorithm}

\subsection{Task Execution}
\label{sec:method_transfer}

After finetuning on the target environment, we combine the GCRL agent $\pi_\tau$, the temporal contrastive mapping $f_w$, the expert demonstration $\tau_{\text{expert}}$, and the cluster classifier to execute tasks. As shown in \autoref{alg:task_execution}, on each step,
we predict the current cluster and select the next sub-goals $g$ as the state that transitions to the next cluster on the shortest path from the current cluster to the target cluster, or the target state if we are already in the target cluster, and execute the action sampled from $\pi_\theta(s, g)$.

\begin{algorithm}[htp!]
\caption{Task Execution}
\label{alg:task_execution}
\begin{algorithmic}[1]
\State \textbf{Input:} env, $\pi_\theta$, $\tau_{\text{expert}}$, $f_w$, ClusterClassifier
\State $s \gets \text{env.reset()}$
\While{not done}
    \State $c \gets \text{ClusterClassifier}(f_w(s))$
    \State $g \gets \text{GetSubGoal}(f_w, c, \tau_{\text{expert}})$
    \State action $ \sim \pi_\theta(s, g)$
    \State $s, \text{ done} \gets \text{env.step(action)}$
\EndWhile
\end{algorithmic}
\end{algorithm}

\begin{figure}[htb!]
    \centering
    \includegraphics[width=0.7\linewidth]{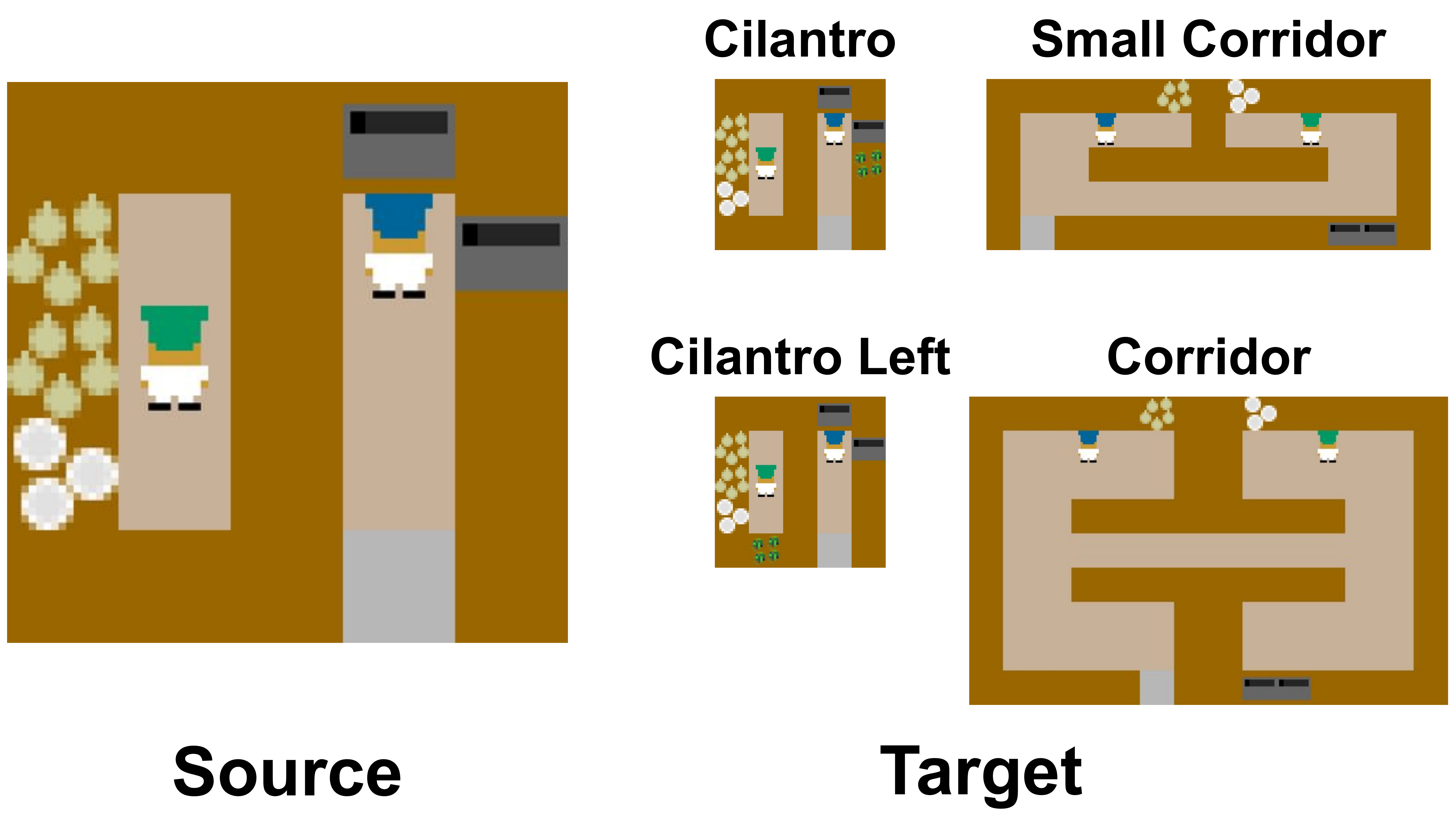}
\caption{
    The source and target Overcooked \cite{carroll2019utility} tasks. The two chefs need to coordinate to make soup and deliver soups. In each environment, there are two chefs (the chef with the green hat and the chef with the blue hat), onion dispensers, plate dispensers, ovens (the grey box with a black top), a serving area (the plain light grey box), walls (brown box) and optionally cilantro dispensers.
}
    \label{fig:env_viz_overcooked}
\end{figure}

\begin{figure}[htb!]
    \centering
    \includegraphics[width=0.8\linewidth]{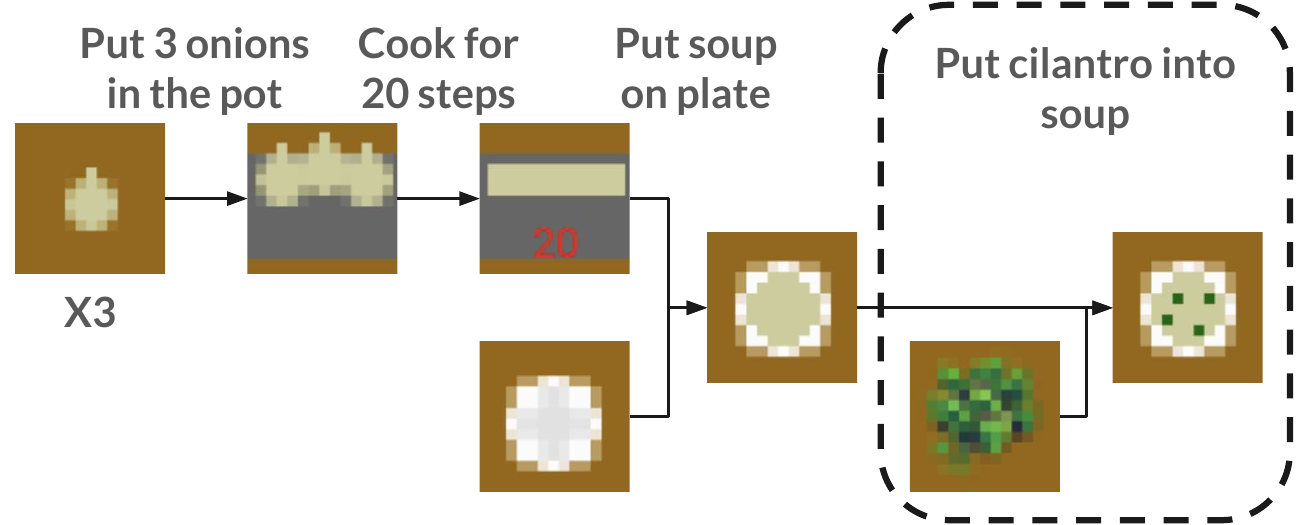}
    \caption{Overcooked recipes.
    To make one soup, the two chefs need to
    1) fetch three onions from the onion dispenser and put them into the oven one by one, and
    2) turn on the oven and wait for 20 steps, and
    3) fetch a plate from the plate dispenser, take the soup from the oven to the plate, and
    4) Optionally, to make a cilantro soup, fetch Cilantro from the dispenser and put it on the soup plate.
    }
    \label{fig:overcooked_recipe}
\end{figure}

\section{Experiments}

In this section, we evaluated our methods for transferring to new environments. We aim to answer the following questions: 
\begin{enumerate*}[label=\arabic*)]
  \item Does our method enable learning in a new environment for single-agent and multi-agents faster?
  \item Are the generated sub-goals qualitatively interpretable?
\end{enumerate*}

\subsection{Setup}

\begin{figure*}[htbp!]
    \centering
    \includegraphics[width=1\linewidth]{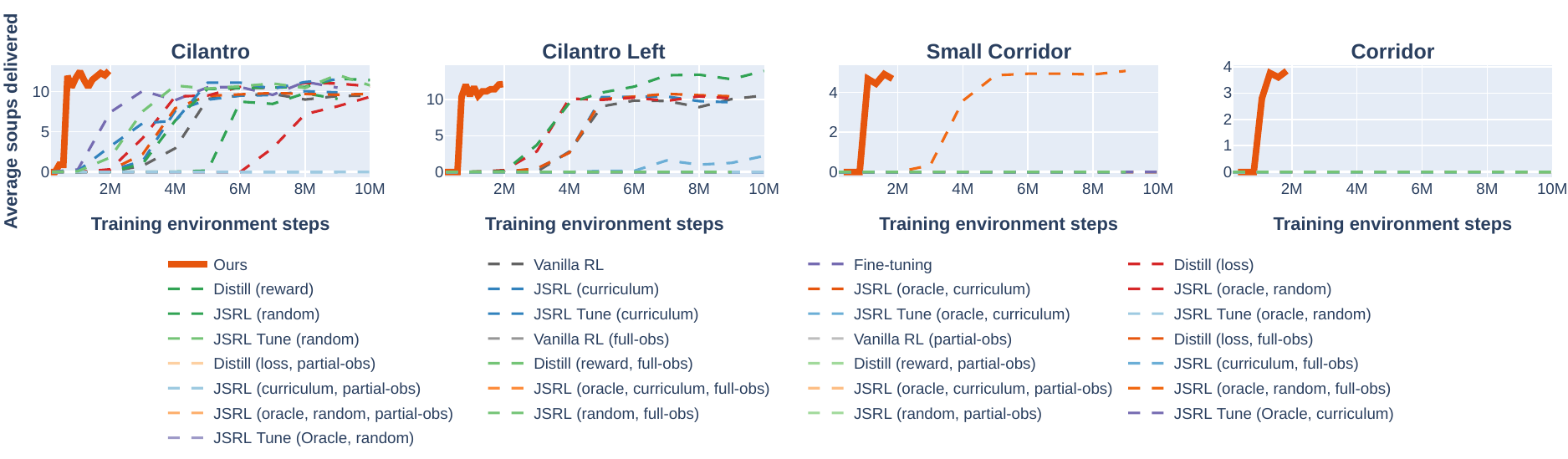}
    \caption{Overcooked Learning Curves. Average soups delivered over 50 episodes throughout training. Most baselines in \textit{small corridor} and \textit{corridor} do not deliver any soups, thus overlapping flat lines.}
    \label{fig:overcooked_transfer_curves}
\end{figure*}

We evaluated our method and five baselines on four transfer learning experiments in the Overcooked environment \cite{carroll2019utility}. Overcooked is a simplified version of the popular video game \textit{Overcooked} \cite{overcooked}, where 2-4 players control chefs cooking and serving dishes in a kitchen. We consider two-player scenarios where the chefs must coordinate to prepare and deliver soups. Each dish recipe contains several high-level steps, as shown in \autoref{fig:overcooked_recipe}.
We pre-trained agents on a source environment $\text{env}_s$ and transferred them to each experiment's target environment $\text{env}_t$. The target environments were designed as variants of the source environment, differing in layout or task. The \textit{Cilantro} and \textit{Cilantro left} environments have different recipes and layouts, and the \textit{small corridor} and \textit{corridor} environments have different layouts. The source and target tasks are visualized in \autoref{fig:env_viz_overcooked}.
We used partially observable agents in all experiments unless specified otherwise. Each episode consisted of 500 timesteps, and the performance was evaluated based on the number of soups delivered per episode. The Overcooked environment has a fixed initial configuration and deterministic dynamics. We randomly rolled out the agent for ten steps before executing the policy to introduce randomness to the environment.
We provide a single expert trajectory for each environment via hard-coded policies.

We compare our method to the following five methods.
\begin{itemize}
    \item \textbf{Vanilla RL}: Training an RL agent from scratch.
    \item \textbf{Fine-tuning}: Fine-tuning the agent trained on the source environment.
    \item \textbf{Policy Distillation (Loss)}: Policy distillation through an auxiliary cross-entropy loss between the action probabilities from the policy pre-trained in the source environment and the learning policy \cite{schmitt2018kickstarting}.
    \item \textbf{Policy Distillation (Reward)}: Policy distillation through a reward shaping term captures the difference of the pre-trained critic in the source environment between current the previous timesteps \cite{czarnecki2019distilling}.
    \item \textbf{JumpStart RL}: JumpStart RL uses a guiding policy to form a curriculum learning, where we gradually sample fewer actions from the guiding policy \cite{uchendu2023jump}. In this paper, we evaluated eight variants of JumpStart RL:
        \begin{enumerate*}[label=\arabic*)]
          \item whether the curriculum schedule is random or specified,
          \item whether the policy is pre-trained on the source environment and
          \item whether the guiding policy is trained on the source or target environment. Note that using policies trained on the target environment as guide policies might give it unfair advantages.
        \end{enumerate*}
\end{itemize}

\subsection{Transfer Learning Results}

\begin{table}[htbp!]
\begin{center}
\resizebox{\linewidth}{!}{
\begin{tabular}{ccccc}
\toprule
\textbf{Environment} & \textbf{Cilantro} & \textbf{Cilantro Left} & \textbf{Small Corridor} & \textbf{Corridor} \\
\midrule
\textbf{Ours} &   \textbf{680.5K} &        \textbf{806.3K} &           \textbf{1.1M} &     \textbf{1.3M} \\
Vanilla RL    &              5.0M &                   6.0M &                     n/a &               n/a \\
Fine-tuning   &              3.0M &                   1.0M &                     n/a &               n/a \\
Distill       &              9.0M &                   3.0M &                     n/a &               n/a \\
JSRL          &              6.2M &                   5.8M &                    5.0M &               n/a \\
\bottomrule
\end{tabular}
}
\end{center}

\caption{Overcooked training steps to convergence (reaching 90\% of the max steps per method per environment) table. n/a means the method did not deliver any soup.}
\label{tab:overcooked_convergence}
\end{table}

\begin{table}[htbp!]
\begin{center}
\resizebox{0.95\linewidth}{!}{
\begin{tabular}{ccccc}
\toprule
\textbf{Environment} & \textbf{Cilantro} & \textbf{Cilantro Left} & \textbf{Small Corridor} & \textbf{Corridor} \\
\midrule
\textbf{Ours} &    \textbf{12.58} &         \textbf{12.10} &           \textbf{4.92} &     \textbf{3.84} \\
Vanilla RL    &              9.72 &                  10.54 &                    0.00 &              0.00 \\
Fine-tuning   &             11.22 &                   0.02 &                    0.00 &              0.00 \\
Distill       &              9.58 &                   0.03 &                    0.00 &              0.00 \\
JSRL          &              8.28 &                   5.97 &                    0.42 &              0.00 \\
\bottomrule
\end{tabular}
}
\end{center}

\caption{Overcooked max soups delivered.}
\label{tab:overcooked_max_soups_delivered}
\end{table}

\begin{figure}
    \centering
    \includegraphics[width=0.9\linewidth]{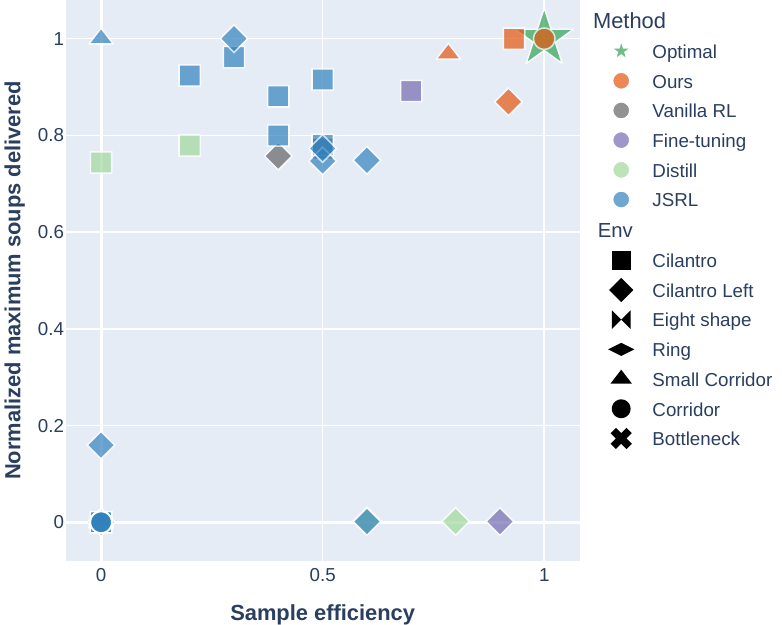}
    \caption{    The scatter plot for normalized performance and sample efficiency in the Overcooked environment.
    The maximum number of soups delivered is normalized using the formula: maximum number of soups delivered for a given method / maximum number of soups delivered for all methods in an environment.
    The Sample efficiency is normalized using the formula: 1 - (steps to convergence for a given method / maximum steps to convergence in the environment). Steps to convergence are determined by the steps at which a method reaches 90\% of its maximum performance. Variants of the same method are grouped under a single plot category.}
    \label{fig:overcooked_comp_scatter}
\end{figure}

We show the average soups delivered for each method throughout training on each environment in \autoref{fig:overcooked_transfer_curves}.
The convergence speed and performance are at \autoref{tab:overcooked_convergence} and \autoref{tab:overcooked_max_soups_delivered}. We compare the normalized convergence speeds and performances across all methods and all environments at \autoref{fig:overcooked_comp_scatter}. Time to convergence is defined as reaching 90\% of maximum performance.
Our methods consistently learn 4.6 times faster on average than the fastest baselines across all experiments, reaching similar or better performances.
On experiments transferring to environments with similar layouts but different tasks from the source environment, the \textit{Cilantro} and \textit{Cilantro left} environments, the transfer learning baselines perform poorly and, sometimes, even worse than the vanilla RL. This is because the guidance from the source environment policy can be biased toward the old behaviors, making it challenging to learn behaviors needed for the new environments. This is especially true in the environments with the cilantro recipes because delivering soups before putting Cilantro in can significantly hinder the resulting performance. This also shows that our method can effectively transfer to environments with tasks where slight differences can result in significant performance degradation.
On experiments transferring to environments with similar tasks but different layouts from the source environment, the \textit{small corridor} and \textit{corridor} environments, our method could effectively transfer to such environments. In contrast, other methods struggle to deliver soups. This is because the narrow and long corridors require agents to coordinate so as not to block others from delivering soups successfully. This demonstrates our method's ability to perform long-horizon multi-agent planning and coordination.

\subsection{Interpretable Sub-goals}

\begin{figure}[htbp]
    \centering
    
    \subfloat[Cilantro]{
        \includegraphics[width=0.9\linewidth]{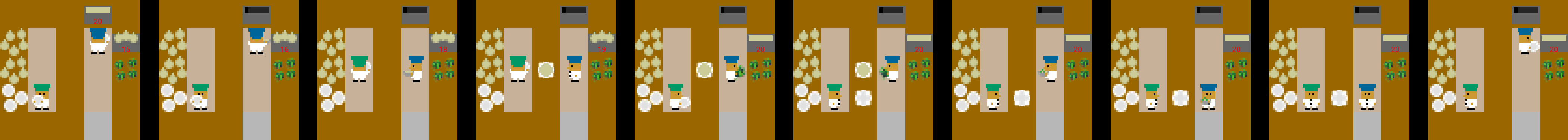}
        \label{fig:overcooked_plan_viz_cilantro}
    }

    \subfloat[Cilantro Left]{
        \includegraphics[width=0.9\linewidth]{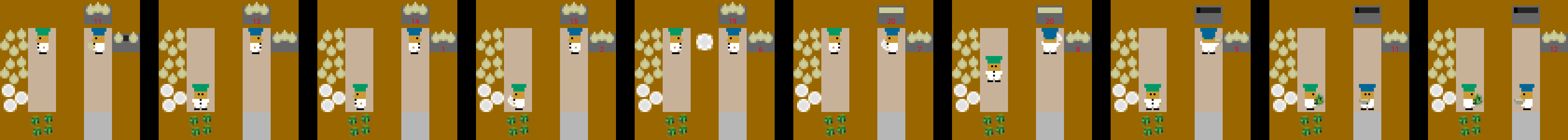}
        \label{fig:overcooked_plan_viz_cilantro_left}
    }

    \subfloat[Small Corridor]{
        \includegraphics[width=0.9\linewidth]{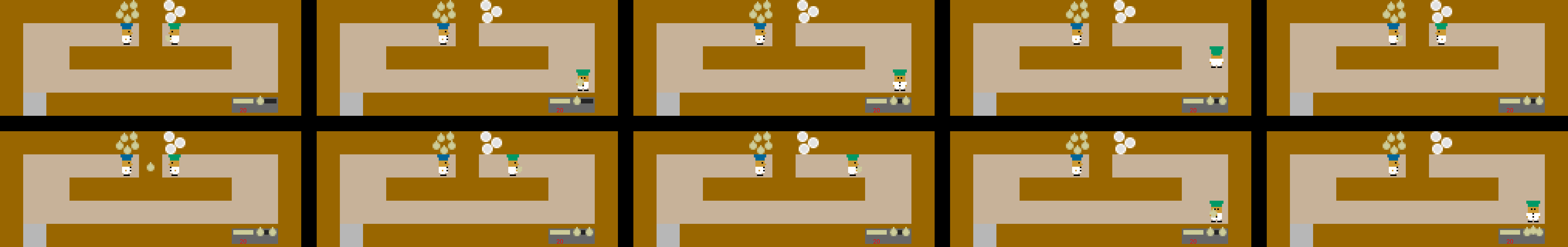}
        \label{fig:overcooked_plan_viz_small_corridor}
    }

    \subfloat[Corridor]{
        \includegraphics[width=0.9\linewidth]{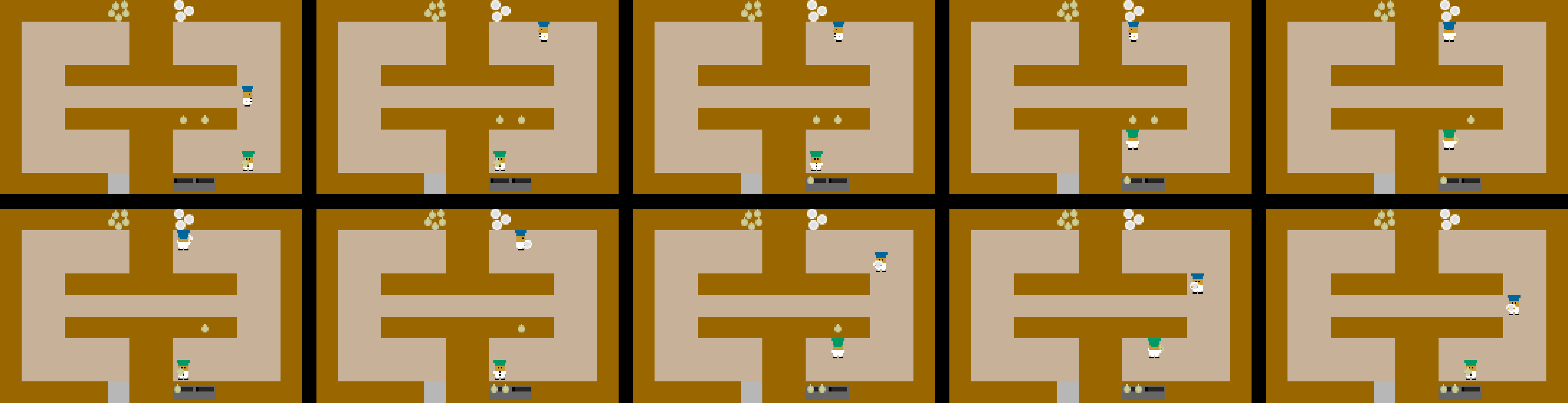}
        \label{fig:overcooked_plan_viz_corridor}
    }

    \caption{Overcooked sub-goals. Samples of sub-goal sequences were generated for each overcooked environment.
    Semantically meaningful breakdown of the task emerges naturally from the temporal contrastive embedding clusters.
    For example, the sub-goals qualitatively demonstrate the intentions for handing over onions, fetching plates, putting onions into the oven, and taking soups out of the oven.
    }
    \label{fig:overcooked_plan_viz}
\end{figure}

The sub-goals generated from \autoref{sec:method_temp_cluster} exhibit semantically meaningful breakdown of tasks, e.g., fetching onions, loading onion to the oven, and serving soups, as shown qualitatively in \autoref{fig:overcooked_plan_viz}. This empirically demonstrated that unsupervised temporal contrastive learning could discover semantically meaningful structures from rollouts. We provide a potential explanation for this. The connection between clusters in the latent space tends to be the connection of a bottleneck structure, where the bottleneck transitions are a sequence of actions that enable the agent to reach previously impossible states. Such transitions often correspond to sub-goals for a task since the agent can advance to previously inaccessible states by following the next sub-goal. One example of such a bottleneck is fetching an onion when the agent has no onion. By fetching an onion, the agent can reach states of carrying onions around that were previously inaccessible.

\section{Conclusion}

This paper introduced a novel transfer learning framework for deep reinforcement learning that combines goal-conditioned policies with unsupervised learning of temporal abstractions. Experiments on Overcooked multi-agent coordination tasks demonstrated the effectiveness of our framework in terms of improved sample efficiency, the ability to solve sparse-reward and long-horizon challenges, and enhanced interpretability through the automatic discovery of meaningful sub-goals. These findings highlight the advantages of integrating goal-conditioned RL with unsupervised temporal abstraction learning for successful transfer to complex target domains, demonstrating superior performance compared to baseline methods such as fine-tuning, policy distillations, and curriculum learning methods.
Compared to state-of-the-art baselines, our method achieves the same or better performances while requiring only 21.7\% of the training samples.
Our work opens up exciting directions for future research, such as integrating language guidance into the contrastive learning process and applying our framework to real-world robotics tasks, paving the way for more intelligent, adaptable, and collaborative AI systems.

\addtolength{\textheight}{-2cm}   


\bibliographystyle{plain}
\bibliography{references}

\begin{thebibliography}{10}

\bibitem{andrychowicz2017hindsight}
Marcin Andrychowicz, Filip Wolski, Alex Ray, Jonas Schneider, Rachel Fong, Peter Welinder, Bob McGrew, Josh Tobin, OpenAI Pieter~Abbeel, and Wojciech Zaremba.
\newblock Hindsight experience replay.
\newblock {\em Advances in neural information processing systems}, 30, 2017.

\bibitem{andrychowicz2020learning}
OpenAI:~Marcin Andrychowicz, Bowen Baker, Maciek Chociej, Rafal Jozefowicz, Bob McGrew, Jakub Pachocki, Arthur Petron, Matthias Plappert, Glenn Powell, Alex Ray, et~al.
\newblock Learning dexterous in-hand manipulation.
\newblock {\em The International Journal of Robotics Research}, 39(1):3--20, 2020.

\bibitem{bacon2017option}
Pierre-Luc Bacon, Jean Harb, and Doina Precup.
\newblock The option-critic architecture.
\newblock In {\em Proceedings of the AAAI conference on artificial intelligence}, volume~31, 2017.

\bibitem{cao2022reinforcement}
Hoang-Giang Cao, Weihao Zeng, and I-Chen Wu.
\newblock Reinforcement learning for picking cluttered general objects with dense object descriptors.
\newblock In {\em 2022 International Conference on Robotics and Automation (ICRA)}, pages 6358--6364. IEEE, 2022.

\bibitem{cao2023learning}
Hoang-Giang Cao, Weihao Zeng, and I-Chen Wu.
\newblock Learning sim-to-real dense object descriptors for robotic manipulation.
\newblock In {\em 2023 IEEE International Conference on Robotics and Automation (ICRA)}, pages 9501--9507. IEEE, 2023.

\bibitem{carroll2019utility}
Micah Carroll, Rohin Shah, Mark~K Ho, Tom Griffiths, Sanjit Seshia, Pieter Abbeel, and Anca Dragan.
\newblock On the utility of learning about humans for human-ai coordination.
\newblock {\em Advances in neural information processing systems}, 32, 2019.

\bibitem{chen2020simple}
Ting Chen, Simon Kornblith, Mohammad Norouzi, and Geoffrey Hinton.
\newblock A simple framework for contrastive learning of visual representations.
\newblock In {\em International conference on machine learning}, pages 1597--1607. PMLR, 2020.

\bibitem{czarnecki2019distilling}
Wojciech~M Czarnecki, Razvan Pascanu, Simon Osindero, Siddhant Jayakumar, Grzegorz Swirszcz, and Max Jaderberg.
\newblock Distilling policy distillation.
\newblock In {\em The 22nd international conference on artificial intelligence and statistics}, pages 1331--1340. PMLR, 2019.

\bibitem{ecoffet2019go}
Adrien Ecoffet, Joost Huizinga, Joel Lehman, Kenneth~O Stanley, and Jeff Clune.
\newblock Go-explore: a new approach for hard-exploration problems.
\newblock {\em arXiv preprint arXiv:1901.10995}, 2019.

\bibitem{finn2017model}
Chelsea Finn, Pieter Abbeel, and Sergey Levine.
\newblock Model-agnostic meta-learning for fast adaptation of deep networks.
\newblock In {\em International conference on machine learning}, pages 1126--1135. PMLR, 2017.

\bibitem{florence2018dense}
Peter~R Florence, Lucas Manuelli, and Russ Tedrake.
\newblock Dense object nets: Learning dense visual object descriptors by and for robotic manipulation.
\newblock {\em arXiv preprint arXiv:1806.08756}, 2018.

\bibitem{frans2017meta}
Kevin Frans, Jonathan Ho, Xi~Chen, Pieter Abbeel, and John Schulman.
\newblock Meta learning shared hierarchies.
\newblock {\em arXiv preprint arXiv:1710.09767}, 2017.

\bibitem{overcooked}
Ghost~Town Games.
\newblock Overcooked.
\newblock \url{https://store.steampowered.com/app/448510/Overcooked/}, 2016.

\bibitem{gupta2017learning}
Abhishek Gupta, Coline Devin, YuXuan Liu, Pieter Abbeel, and Sergey Levine.
\newblock Learning invariant feature spaces to transfer skills with reinforcement learning.
\newblock {\em arXiv preprint arXiv:1703.02949}, 2017.

\bibitem{higgins2017darla}
Irina Higgins, Arka Pal, Andrei Rusu, Loic Matthey, Christopher Burgess, Alexander Pritzel, Matthew Botvinick, Charles Blundell, and Alexander Lerchner.
\newblock Darla: Improving zero-shot transfer in reinforcement learning.
\newblock In {\em International Conference on Machine Learning}, pages 1480--1490. PMLR, 2017.

\bibitem{hu2023planning}
Edward~S Hu, Richard Chang, Oleh Rybkin, and Dinesh Jayaraman.
\newblock Planning goals for exploration.
\newblock {\em arXiv preprint arXiv:2303.13002}, 2023.

\bibitem{laskin2020curl}
Michael Laskin, Aravind Srinivas, and Pieter Abbeel.
\newblock Curl: Contrastive unsupervised representations for reinforcement learning.
\newblock In {\em International conference on machine learning}, pages 5639--5650. PMLR, 2020.

\bibitem{laskin2020reinforcement}
Misha Laskin, Kimin Lee, Adam Stooke, Lerrel Pinto, Pieter Abbeel, and Aravind Srinivas.
\newblock Reinforcement learning with augmented data.
\newblock {\em Advances in neural information processing systems}, 33:19884--19895, 2020.

\bibitem{levy2017learning}
Andrew Levy, George Konidaris, Robert Platt, and Kate Saenko.
\newblock Learning multi-level hierarchies with hindsight.
\newblock {\em arXiv preprint arXiv:1712.00948}, 2017.

\bibitem{li2021genurl}
Siyuan Li, Zicheng Liu, Zelin Zang, Di~Wu, Zhiyuan Chen, and Stan~Z Li.
\newblock Genurl: A general framework for unsupervised representation learning.
\newblock {\em arXiv preprint arXiv:2110.14553}, 2021.

\bibitem{lu2021aw}
Yao Lu, Karol Hausman, Yevgen Chebotar, Mengyuan Yan, Eric Jang, Alexander Herzog, Ted Xiao, Alex Irpan, Mohi Khansari, Dmitry Kalashnikov, et~al.
\newblock Aw-opt: Learning robotic skills with imitation and reinforcement at scale.
\newblock {\em arXiv preprint arXiv:2111.05424}, 2021.

\bibitem{ma2023eureka}
Yecheng~Jason Ma, William Liang, Guanzhi Wang, De-An Huang, Osbert Bastani, Dinesh Jayaraman, Yuke Zhu, Linxi Fan, and Anima Anandkumar.
\newblock Eureka: Human-level reward design via coding large language models.
\newblock {\em arXiv preprint arXiv:2310.12931}, 2023.

\bibitem{mendonca2021discovering}
Russell Mendonca, Oleh Rybkin, Kostas Daniilidis, Danijar Hafner, and Deepak Pathak.
\newblock Discovering and achieving goals via world models.
\newblock {\em Advances in Neural Information Processing Systems}, 34:24379--24391, 2021.

\bibitem{nachum2018data}
Ofir Nachum, Shixiang~Shane Gu, Honglak Lee, and Sergey Levine.
\newblock Data-efficient hierarchical reinforcement learning.
\newblock {\em Advances in neural information processing systems}, 31, 2018.

\bibitem{oord2018representation}
Aaron van~den Oord, Yazhe Li, and Oriol Vinyals.
\newblock Representation learning with contrastive predictive coding.
\newblock {\em arXiv preprint arXiv:1807.03748}, 2018.

\bibitem{park2024foundation}
Seohong Park, Tobias Kreiman, and Sergey Levine.
\newblock Foundation policies with hilbert representations.
\newblock {\em arXiv preprint arXiv:2402.15567}, 2024.

\bibitem{plappert2018multi}
Matthias Plappert, Marcin Andrychowicz, Alex Ray, Bob McGrew, Bowen Baker, Glenn Powell, Jonas Schneider, Josh Tobin, Maciek Chociej, Peter Welinder, et~al.
\newblock Multi-goal reinforcement learning: Challenging robotics environments and request for research.
\newblock {\em arXiv preprint arXiv:1802.09464}, 2018.

\bibitem{pong2019skew}
Vitchyr~H Pong, Murtaza Dalal, Steven Lin, Ashvin Nair, Shikhar Bahl, and Sergey Levine.
\newblock Skew-fit: State-covering self-supervised reinforcement learning.
\newblock {\em arXiv preprint arXiv:1903.03698}, 2019.

\bibitem{rusu2016progressive}
Andrei~A Rusu, Neil~C Rabinowitz, Guillaume Desjardins, Hubert Soyer, James Kirkpatrick, Koray Kavukcuoglu, Razvan Pascanu, and Raia Hadsell.
\newblock Progressive neural networks.
\newblock {\em arXiv preprint arXiv:1606.04671}, 2016.

\bibitem{schaul2015universal}
Tom Schaul, Daniel Horgan, Karol Gregor, and David Silver.
\newblock Universal value function approximators.
\newblock In {\em International conference on machine learning}, pages 1312--1320. PMLR, 2015.

\bibitem{schmitt2018kickstarting}
Simon Schmitt, Jonathan~J Hudson, Augustin Zidek, Simon Osindero, Carl Doersch, Wojciech~M Czarnecki, Joel~Z Leibo, Heinrich Kuttler, Andrew Zisserman, Karen Simonyan, et~al.
\newblock Kickstarting deep reinforcement learning.
\newblock {\em arXiv preprint arXiv:1803.03835}, 2018.

\bibitem{schulman2017proximal}
John Schulman, Filip Wolski, Prafulla Dhariwal, Alec Radford, and Oleg Klimov.
\newblock Proximal policy optimization algorithms.
\newblock {\em arXiv preprint arXiv:1707.06347}, 2017.

\bibitem{sekar2020planning}
Ramanan Sekar, Oleh Rybkin, Kostas Daniilidis, Pieter Abbeel, Danijar Hafner, and Deepak Pathak.
\newblock Planning to explore via self-supervised world models.
\newblock In {\em International conference on machine learning}, pages 8583--8592. PMLR, 2020.

\bibitem{silver2017mastering}
David Silver, Thomas Hubert, Julian Schrittwieser, Ioannis Antonoglou, Matthew Lai, Arthur Guez, Marc Lanctot, Laurent Sifre, Dharshan Kumaran, Thore Graepel, et~al.
\newblock Mastering chess and shogi by self-play with a general reinforcement learning algorithm.
\newblock {\em arXiv preprint arXiv:1712.01815}, 2017.

\bibitem{sutton1999between}
Richard~S Sutton, Doina Precup, and Satinder Singh.
\newblock Between mdps and semi-mdps: A framework for temporal abstraction in reinforcement learning.
\newblock {\em Artificial intelligence}, 112(1-2):181--211, 1999.

\bibitem{uchendu2023jump}
Ikechukwu Uchendu, Ted Xiao, Yao Lu, Banghua Zhu, Mengyuan Yan, Jos{\'e}phine Simon, Matthew Bennice, Chuyuan Fu, Cong Ma, Jiantao Jiao, et~al.
\newblock Jump-start reinforcement learning.
\newblock In {\em International Conference on Machine Learning}, pages 34556--34583. PMLR, 2023.

\bibitem{vemula2023virtues}
Anirudh Vemula, Yuda Song, Aarti Singh, Drew Bagnell, and Sanjiban Choudhury.
\newblock The virtues of laziness in model-based rl: A unified objective and algorithms.
\newblock In {\em International Conference on Machine Learning}, pages 34978--35005. PMLR, 2023.

\bibitem{vinyals2017starcraft}
Oriol Vinyals, Timo Ewalds, Sergey Bartunov, Petko Georgiev, Alexander~Sasha Vezhnevets, Michelle Yeo, Alireza Makhzani, Heinrich K{\"u}ttler, John Agapiou, Julian Schrittwieser, et~al.
\newblock Starcraft ii: A new challenge for reinforcement learning.
\newblock {\em arXiv preprint arXiv:1708.04782}, 2017.

\bibitem{zhan2022learning}
Albert Zhan, Ruihan Zhao, Lerrel Pinto, Pieter Abbeel, and Michael Laskin.
\newblock Learning visual robotic control efficiently with contrastive pre-training and data augmentation.
\newblock In {\em 2022 IEEE/RSJ International Conference on Intelligent Robots and Systems (IROS)}, pages 4040--4047. IEEE, 2022.

\bibitem{zhang2017deep}
Jingwei Zhang, Jost~Tobias Springenberg, Joschka Boedecker, and Wolfram Burgard.
\newblock Deep reinforcement learning with successor features for navigation across similar environments.
\newblock In {\em 2017 IEEE/RSJ International Conference on Intelligent Robots and Systems (IROS)}, pages 2371--2378. IEEE, 2017.

\bibitem{zhu2023transfer}
Zhuangdi Zhu, Kaixiang Lin, Anil~K Jain, and Jiayu Zhou.
\newblock Transfer learning in deep reinforcement learning: A survey.
\newblock {\em IEEE Transactions on Pattern Analysis and Machine Intelligence}, 2023.

\end{thebibliography}

\end{document}